\documentclass[11pt,letterpaper]{article}
\usepackage{cogsys}

\usepackage[T1]{fontenc}
\usepackage{times}
\usepackage[pdftex]{graphicx} 
\usepackage[colorinlistoftodos]{todonotes}
\usepackage{inconsolata}
\usepackage{booktabs}
\usepackage[normalem]{ulem}

\usepackage[symbol]{footmisc}
\usepackage[font=small,labelfont=bf,tableposition=top]{caption}

\usepackage[margins]{trackchanges}
\addeditor{SM}
\addeditor{MK}

\usepackage{enumitem}
\setlist{nolistsep}
\setlist{nosep}

\renewcommand{\thefootnote}{\fnsymbol{footnote}}

\usepackage{natbib}
\begin{document}

\title{Characterizing an Analogical Concept Memory for Architectures Implementing the Common Model of Cognition}
\author{Shiwali Mohan}{shiwali.mohan@parc.com}
\author{Matt Klenk}{klenk@parc.com}
\author{Matthew Shreve}{matthew.shreve@parc.com}
\author{Kent Evans}{kent.evans@parc.com}
\author{Aaron Ang}{aaron.ang@parc.com}
\author{John Maxwell}{john.maxwell@parc.com}
\address{Palo Alto Research Center,  Palo Alto, CA 94306 USA}
\vskip 0.2in

\begin{abstract}
Architectures that implement the Common Model of Cognition - Soar, ACT-R, and Sigma - have a prominent place in research on cognitive modeling as well as on designing complex intelligent agents. In this paper, we explore how computational models of analogical processing can be brought into these architectures to enable concept acquisition from examples obtained interactively. We propose a new analogical concept memory for Soar that augments its current system of declarative long-term memories. We frame the problem of concept learning as embedded within the larger context of interactive task learning (ITL) and embodied language processing (ELP). We demonstrate that the analogical learning methods implemented in the proposed memory can quickly learn a diverse types of novel concepts that are useful not only in recognition of a concept in the environment but also in action selection. Our approach has been instantiated in an implemented cognitive system \textsc{Aileen} and evaluated on a simulated robotic domain.
\end{abstract}

\section{Introduction}
The recent proposal for the common model of cognition (CMC; \citeauthor{laird2017standard} \citeyear{laird2017standard}) identifies the central themes in the past $30$ years of research in three cognitive architectures - Soar \citep{laird2012}, ACT-R \citep{anderson2009can}, and Sigma \citep{rosenbloom2016sigma}. These architectures have been prominent not only in cognitive modeling but also in designing complex intelligent agents. CMC architectures aim to implement a set of domain-general computational processes which operate over domain-specific knowledge to produce effective task behavior. Early research in CMC architectures studied procedural knowledge - the knowledge of \emph{how} to perform tasks, often expressed as \emph{if-else} rules. It explored the computational underpinnings of a general purpose decision making process that can apply hand-engineered procedural knowledge to perform a wide-range of tasks. Later research studied various ways in which procedural knowledge can be learned and optimized.

While CMC architectures have been applied widely, \cite{hinrichs2017towards} note that reasoning in them focuses exclusively on problem solving, decision making, and behavior. Further, they argue that a distinctive and arguably signature feature of human intelligence is being able to build complex conceptual structures of the world. In the CMC terminology, the knowledge of concepts is \emph{declarative} knowledge - the knowledge of \emph{what}. An example of declarative knowledge is the final goal state of the tower-of-hanoi puzzle. Procedural knowledge in tower-of-hanoi are the set of rules that guide action selection in service of achieving the goal state. CMC architectures agree that conceptual structures are useful for intelligent behavior. For instance, to solve tower-of-hanoi, understanding the goal state is critical. However, there is limited understanding of how declarative knowledge about the world is acquired in CMC architectures.

In this paper, we study the questions of declarative concept representation, acquisition, and usage in task performance in a prominent CMC architecture - Soar. As it bears significant similarities with ACT-R and Sigma in the organization of computation and information, our findings can be generalized to those architectures as well. Soar has two declarative long-term memories - episodic and semantic - each with its own distinctive function. Episodic memory aims at providing access to past contextual experiences. Semantic memory is aimed at representing and providing access to general factual knowledge about the world. We augment this system with a new \emph{concept} memory that aims at acquiring general knowledge about the world by collecting and analyzing similar experiences, functionally bridging the episodic and semantic memories.

To design the concept memory, we leverage the computational processes that underlie analogical reasoning and generalization in the Companions cognitive architecture - the Structure Mapping Engine (SME; \citeauthor{forbus2017extending} \citeyear{forbus2017extending}) and the Sequential Analogical Generalization Engine (SAGE; \citeauthor{mclure2015extending} \citeyear{mclure2015extending}). Analogical matching, retrieval, and generalization is the foundation of the Companions Cognitive architecture. In \textit{Why we are so smart?}, Gentner claims that what makes human cognition superior to other animals is ``First, relational concepts are critical to higher-order cognition, but relational concepts are both non-obvious in initial learning and elusive in memory retrieval. Second, analogy is the mechanism by which relational knowledge is revealed. Third, language serves both to invite learning relational concepts and to provide cognitive stability once they are learned'' \citep{gentner2003we}. Gentner's observations provide a compelling case for exploring analogical processing as a basis for concept learning.

Our design exploration is motivated by the interactive task learning problem (ITL; \citeauthor{gluck2019interactive} \citeyear{gluck2019interactive}) in embodied agents. ITL agents rely on natural interaction modalities such as embodied dialog to learn new tasks. Conceptual knowledge and language are inextricably tied - language is a medium through which conceptual knowledge about the world is communicated and learned. Consequently, embodied language processing (ELP) for ITL provides a set of functional requirements that an architectural concept memory must address. Embedding concept learning within the ITL and ELP contexts is a significant step forward from previous explorations in concept formation. Prior approaches have studied concept formation independently of how they will be used in a complex cognitive system, often focusing on the problems of recognizing the existence of a concept in input data and organizing concepts into a similarity-based hierarchy.

The contributions of this paper are follows. It,
\begin{enumerate}
\item defines the concept formation problem as embedded within the larger context of ELP and ITL;
\item identifies a desiderata for an architectural concept memory;
\item implements a concept memory for Soar agents using the models of analogical processing;
\item introduces a novel interactive process - curriculum of guided participation - that is used for online concept learning;
\item introduces a novel framework for evaluating interactive concept formation in cognitive systems.
\end{enumerate}
Our implementation is a functional (and not an architectural) integration of analogical processing in Soar's declarative long-term memory systems. It characterizes how an analogical concept memory can be interfaced with the current mechanisms. Through experiments and system demonstration, we show that an analogical concept memory:
\begin{enumerate}
    \item supports learning of diverse types of concepts useful in ITL;
    \item supports recognition of concepts for embodied language processing as well as acting based on those concepts;
    \item learns from few examples provided interactively.
\end{enumerate}

\section{Preliminaries}
\label{sec:preliminaries}
\paragraph{Declarative Long-Term Memories in Soar}
In the past two decades, algorithmic research in Soar has augmented the architecture with decalartive long-term memories (dLTMs). Soar has two - semantic  \citep{derbinsky2010towards} and episodic \citep{derbinsky2009efficiently} - that serve distinct cognitive functions following the hypotheses about organization of memory in humans \citep{tulving2005oxford}. Semantic memory enables enriching what is currently observed in the world with what is known generally about it. For example, if a dog is observed in the environment, for certain types of tasks it may be useful to elaborate that it is a type of a mammal. Episodic memory gives an agent a personal history which can later be recalled to establish reference to shared experience with a collaborator, to aid in decision-making by predicting the outcome of possible courses of action, to aid in reasoning by creating an internal model of the environment, and by keeping track of progress on long-term goals. The history is also useful in deliberate reflection about past events to improve behavior through other types of learning such as reinforcement learning or explanation-based learning. Using dLTMs in Soar agents has enable reasoning complexity that wasn't possible earlier \citep{xu2010instance,mohan2014learning,kirk2014interactive, mininger2018interactively}.

However, a crucial question remains unanswered - how is conceptual knowledge in semantic memory acquired? We posit that conceptual knowledge is acquired in two distinctive ways. \citet{kirk2014interactive, kirk2019learning} explore the view that semantic knowledge is acquired through interactive instruction when natural language describes relevant declarative knowledge. An example concept is the goal of tower-of-hanoi \emph{a small block is on a  medium block and a large block is below the medium block.} Here, the trainer provides the definition of the concept declaratively which is later operationalized so that it can be applied to recognize existence and in applying actions while solving tower-of-hanoi. We explore a different view that that conceptual knowledge is acquired through examples demonstrated as a part of instruction. The input from the trainer help group concrete experiences together and a generalization process distills common elements to form a concept definition.

\paragraph{Analogical Processing}
Our approach builds on the analogical concept learning work done in Companions \citep{hinrichs2017towards}. Previous analogical learning work includes spatial prepositions \cite{lockwood2009using}, spatial concepts \citep{mclure2015extending}, physical reasoning problems \citep{klenk2011using},  and activity recognition \citep{chen2019human}. This diversity of reasoning tasks motivates our use of analogical processing to develop an architectural concept memory. Adding to this line of research, our work shows that you can learn a variety of conceptual knowledge within a single system. Furthermore, that such a system can be applied to not only learn how to recognize the concepts but also acting on them in the environment within an interactive task learning session.

\paragraph{Embodied Language Processing and ITL}
Language is an immensely powerful communication medium for humans enabling exchange of information pertaining to not only current events but to events that have happened in the past, events that may happen in the future, and events that are unlikely to have ever occurred. To make progress towards generally intelligent agents that can assist and collaborate with humans, a capability to effectively use language is crucial. We focus on language capabilities necessary to achieve collaboration in physical worlds and term this capability embodied language processing (ELP). ELP can enable novel learning capabilities such as ITL in artificial agents. The main goal of ELP is to establish reference to entities and actions in the shared space such that joint goals can be achieved. Consider the scene in Figure \ref{fig:architecture} in which a human collaborator asks the robot to \emph{move the blue cone to the left of red cylinder}. For successful collaboration, the robot must localize \emph{blue cone} as a specific object on the scene and execute an action on that object. The robot must solve the inverse problem to generate language. It must be able to describe objects and actions that are relevant to its own task performance such that the human collaborator is able to correctly identify them. This paper builds upon a method for ELP - the Indexical Model of situated comprehension \citep{Mohan2014indexical} which is a computational, architectural instantiation of the \emph{Indexical Hypothesis} \citep{glenberg1999indexical}. This model has been shown to be useful in ITL \citep{mohan2014learning, kirk2014interactive}.

\section{\textsc{Aileen}}
\label{sec:aileen}
\begin{figure}
    \centering
    \includegraphics[width=0.9\textwidth]{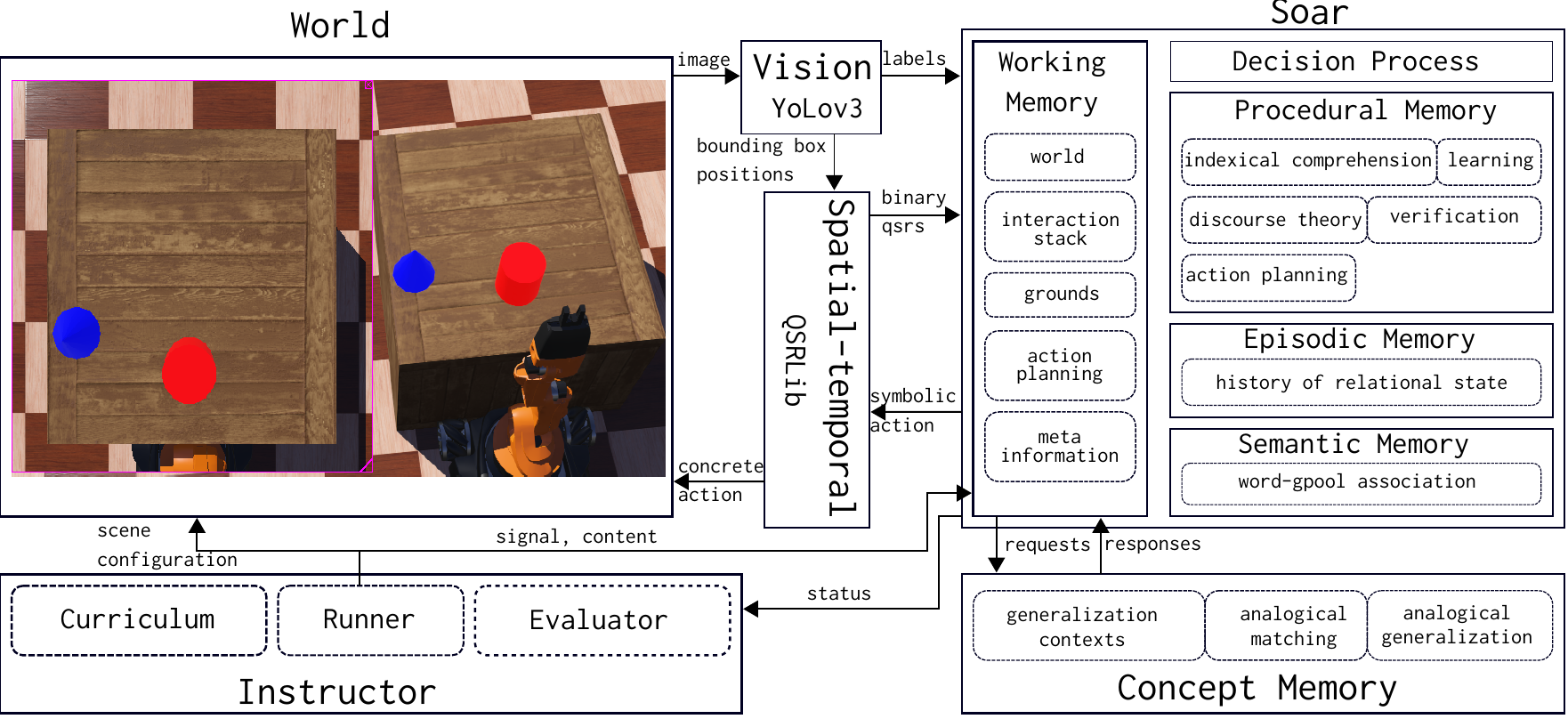}
    \caption{System diagram for Advanced cognItive LEarning for Embodied compreheNsion (\textsc{Aileen})}
    \label{fig:architecture}
\end{figure}

\textsc{Aileen} is a cognitive system that learns new concepts through interactive experiences with a trainer in a simulated world. A system diagram is shown in Figure \ref{fig:architecture}. \textsc{Aileen} lives in a simulated robotic world built in Webots\footnote{https://www.cyberbotics.com/}. The world contains a table-top on which various simple objects can be placed. A simulated camera above the table captures top-down visual information. \textsc{Aileen} is engaged in a continuous \emph{perceive-decide-act} loop with the world. A trainer can set up a scene in the simulated world by placing simple objects on the scene and providing instructions to the agent. \textsc{Aileen} is designed in Soar which has been integrated with a deep learning-based vision module and an analogical concept memory. It is related to \textsc{Rosie}, a cognitive system that has demonstrated interactive, flexible learning on a variety of tasks \citep{mohan2012acquiring,Mohan2014indexical,mohan2014learning,kirk2014interactive,mininger2018interactively}, and implements a similar organization of knowledge.

\paragraph{Visual Module}
The visual module processes the image taken from the simulated camera. It produces output in two channels: object detections as bounding boxes whose centroids are localized on the table-top and two perceptual symbols or \emph{percept}s corresponding to the object's shape and color each. The module is built using a deep learning framework - You Only Look Once (\textsc{YoLo}: \citet{redmon2016you}). \textsc{YoLo} is pre-trained with supervision from the ground truth in the simulator ($12,000$ images). It is detects four shapes (error rate $< 0.1\%$) - \emph{box} (percept - \texttt{CVBox}), \emph{cone} (\texttt{CVCone}), \emph{ball} (\texttt{CVSphere}), and \emph{cylinder} (\texttt{CVCylinder}).

For colors, each detected region containing an object is cropped from the image, and a $K$-means clustering is applied all color pixel values within the crop. Next, two weighted heuristics are applied that selects the cluster that likely comprises the detected shape among any background pixels and/or neighboring objects. The first heuristic selects the cluster with the maximum number of pixels. The second heuristic selects the cluster with the centroid that is closest to the image center of the cropped region. The relative weighted importance of each of these heuristics is then trained using a simple grid search over $w_1$ and $w_2$: $Score = w_1R_s+w_2(1-C_s), s\in D$, where $w_1+w_2=1$, $D$ is the set clusters, $R_s$ denotes the ratio between the number of pixels in each cluster and the the number of pixels in the image crop, and $C_s$ is the Euclidean distance between the centroid of the cluster and the image center normalized by the cropped image width. The average \textsc{RGB} value for all pixels included in the cluster with the highest score is calculated and compared with the preset list of color values. The color label associated with the color value that has the smallest Euclidean distance to the average RGB value is selected. The module can recognize $5$ colors (error rate $< 0.1\%$): \texttt{CVGreen}, \texttt{CVBlue}, \texttt{CVRed}, \texttt{CVYellow}, and \texttt{CVPurple}. Note that the percepts are named so to be readable for system designers - the agent does not rely on the percept symbol strings for any reasoning.

\paragraph{Spatial Processing Module}
The spatial processing module uses \textsc{QSRLib} \citep{gatsoulis2016qsrlib} to process the bounding boxes and centroids generated by the visual module to generate a qualitative description of the spatial configuration of objects. For every pair of objects, the module extracts qualitative descriptions using two spatial calculi (qsrs): cardinal direction (CDC) and region connection (RCC8). Additionally, the spatial processing module can also convert a set of calculi into regions and sample points from them. This enables \textsc{Aileen} to identify locations in continuous space that satisfy qualitative spatial constraints when planning actions.

\paragraph{Intrinsic and Extrinsic Behaviors}
The outputs of the visual module and the spatial module are compiled together to form an object-oriented relational representation of the current state of the world. This representation is written to Soar's working memory graph. Procedural knowledge is encoded as rules in Soar and similarly to \textsc{Rosie} \citep{mohan2012acquiring} consists of knowledge for:
\begin{enumerate}
    \item Interaction: As in \textsc{Rosie} \citep{mohan2012acquiring} \textsc{Aileen} implements collaborative discourse theory \citep{rich2001collagen} to manage its interactive behavior. It captures the state of task-oriented interaction and is integrated with comprehension, task execution, and learning.
    \item Comprehension: \textsc{Aileen} implements the Indexical Model of comprehension \citep{Mohan2014indexical} to process language by grounding it in the world and domain knowledge. This model formulates language understanding as a search process. It interprets linguistic symbols and their associated semantics as cues to search the current environment as well as domain knowledge. Formulating language comprehension in this fashion integrates naturally with interaction and learning where ambiguities and failures in the search process drive interaction and learning.
    \item External task execution: \textsc{Aileen} has been programmed with primitive actions that enable it to manipulate its environment: \texttt{point(o)}, \texttt{pick-up(o)}, and \texttt{place([x, y, z])}. Following \citet{mohan2014learning}, each primitive action has a proposal rule that encodes its pre-conditions, a model that captures state changes expected to occur when the action is applied, and an application rule. Additionally, given a task goal, \textsc{Aileen} can use iterative-deepening search to plan a sequence of primitive actions to achieve the goal and execute the task in the world.
    \item Learning: Learning in \textsc{Aileen} is the focus of this paper and is significantly different from \textsc{Rosie}. \textsc{Rosie} uses an interactive variation of explanation-based learning \citep{mohan2014learning} to learn representation and execution of tasks.  \textsc{Aileen} uses analogical reasoning and generalization to learn diverse concepts including those relevant to task performance (Sections \ref{sec:icl} and \ref{sec:concept-memory}). A crucial distinction is that EBL requires a complete domain theory to correctly generalize observed examples while analogical reasoning and generalization can operate with partial domain theory by leveraging statistical information in observed examples.
\end{enumerate}
The ongoing ITL research in Soar demonstrates the strength of this organization of knowledge in hybrid cognitive systems. Our conjecture is that an ideal concept memory in an architecture must support complex, integrated, intelligent behavior such as ELP and ITL.

\section{The Interactive Concept Learning Problem}
\label{sec:icl}
To facilitate interactive task learning in cognitive systems like \textsc{Aileen}, concept representation must support: (1) indexical comprehension, which processes language to bring to attention various parts of the environment and domain knowledge and (2) task execution, which applies a sequence of actions to achieve relevant goals in the world. In Section \ref{sec:concepts-in-aileen}, we describe how \textsc{Aileen} uses its conceptual knowledge to drive these two behaviors. Then, in Section \ref{sec:concept-space}, we discuss how analogical processing can be applied to learn conceptual knowledge. We introduce a novel interactive process to teach \textsc{Aileen} new concepts online in Section \ref{sec:curriculum}. And finally, in Section \ref{sec:desiderata}, we delineate the desiderata for a concept memory that supports concept acquisition in \textsc{Aileen} such that these concepts can be used to drive interactive behavior as well as execute tasks.

\subsection{Using Conceptual Knowledge in \textsc{Aileen}}
\label{sec:concepts-in-aileen}
Consider the world in Figure \ref{fig:architecture} and the corresponding working memory graph in Figure \ref{fig:reperesntation-indexical}. Semantic memory stores concept definitions corresponding to various words used to interact with \textsc{Aileen}.  \emph{Maps} (\texttt{M1}, \texttt{M2}, \texttt{M3}, \texttt{M4}, \texttt{M5},texttt{M6}) -  in semantic memory (shown in Figure \ref{fig:reperesntation-indexical}) - associate words (\emph{cylinder}) to their conceptual definition (\texttt{percept CVCylinder}). Maps provide bi-directional access to the association between words and concept definitions. The semantic memory can be queried with a word to retrieve its concept definition. The semantic memory can also we queried with a concept definition to access the word that describes it.

Phrases (1) \emph{blue cone left of red cylinder} and (2) \emph{move blue cone right of red cylinder} can be understood via indexical comprehension (details by as follows:

\begin{figure}
    \centering
    \includegraphics{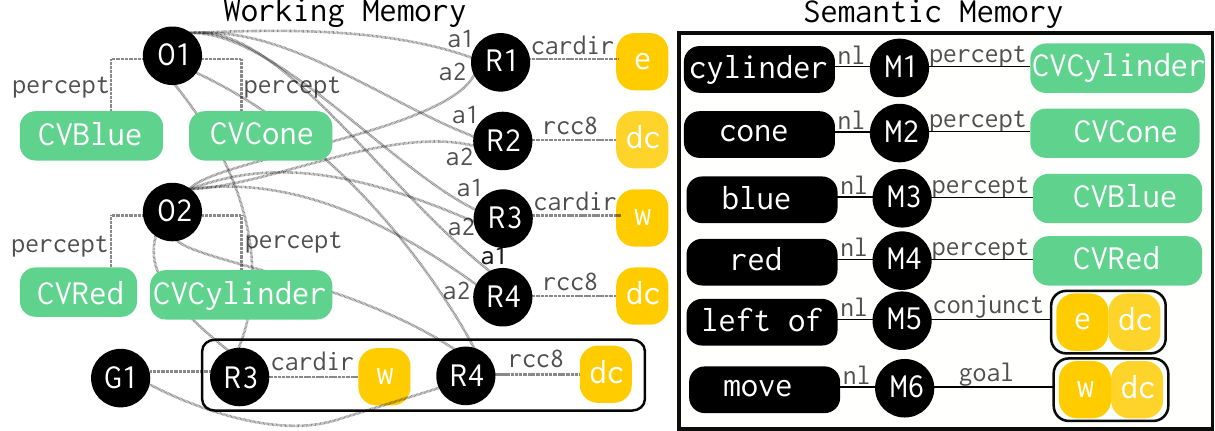}
    \caption{(left) Simplified, partial working memory graph for the scene in Figure \ref{fig:architecture}. Green colored symbols are generated in the visual module and yellow colored symbols are generated in the spatial module. Black colored symbols are internal to Soar and are used for driving behavior. (right) Concepts in semantic memory. Map nodes (\texttt{M1}, \texttt{M2}, \texttt{M3}, \texttt{M4}, \texttt{M5}, \texttt{M6}) connect words with their conceptual definitions in semantic memory.}
    \label{fig:reperesntation-indexical}
\end{figure}

\begin{enumerate}
    \item \emph{Parse the linguistic input into semantic components}. Both (1) and (2) have two references to objects: \texttt{\{or1: obj-ref\{property:blue, property:cone\}\}} and \texttt{\{or2: obj-ref \{property:red, property: cylinder\}\}}. Additionally, (1) has a reference to a spatial relationship: \texttt{\{rel1: \{rel-name: left of, argument1: or1, argument2: or2\}\}}. (2) has a reference to an action: \texttt{\{act1: \{act-name: move, argument1: or1, argument2: or2, relation: left of\} \}}. For this paper, we assume that the knowledge for this step is pre-encoded.

    \item \emph{Create a goal for grounding each reference}. The goal of processing an object reference is to find a set of objects that satisfy the properties specified. It starts with first resolving properties. The process queries semantic memory for a percept that corresponds to various properties in the parse. If the knowledge in Figure \ref{fig:reperesntation-indexical} is assumed, property \texttt{blue} resolves to percept \texttt{CVBlue}, \texttt{cone} to \texttt{CVCone}, \texttt{red} to \texttt{CVRed}, and \texttt{cylinder} to \texttt{CVCylinder}. Using these percepts, \textsc{Aileen} queries its scene to resolve object references. For \texttt{or1}, it finds an object that has both \texttt{CVBlue} and \texttt{CVCone} in its description. Let \texttt{or1} resolve to \texttt{o1} and \texttt{or2} to \texttt{o2} where \texttt{o1} and \texttt{o1} are identifiers of objects visible on the scene. The goal of processing a relation reference is to find a set of spatial calculi that correspond to the name specified. If knowledge in Figure \ref{fig:reperesntation-indexical} is assumed, \texttt{rel1} in (1) is resolved to a conjunction of qsrs \texttt{e(a1,a2)}$\land$\texttt{dc(a1,a2)} i.e, object mapping to \texttt{a1} should be east (in CDC) of \texttt{a2} and they should be disconnected. Similarly, \texttt{act1} in (2) resolves to a task goal which is a conjunction of qsrs \texttt{w(a1,a2)}$\land$\texttt{dc(a1,a2)}

    \item \emph{Compose all references}: Use semantic constraints to resolve the full input. For (1) and (2) \texttt{a1} is matched to to \texttt{ar1} and consequently to \texttt{o1}. Similarly, \texttt{a2} is resolved to \texttt{o2} via \texttt{ar2}.
\end{enumerate}

Tasks are represented in \textsc{Aileen} as goals that it must achieve in its environment. Upon being asked to execute a task, \emph{move blue cone right of red cylinder}, indexical comprehesion determines the desired goal state as \texttt{w(a1,a2)}$\land$\texttt{dc(a1,a2)}. Now, \textsc{Aileen} must execute a sequence of actions to achieve this desired goal state in its environment. Leveraging standard pre-conditions and effects of actions, \textsc{Aileen} can simulate the results of applying plausible actions in any state. Through an iterative deepening search conducted over actions, \textsc{Aileen} can generate and execute a plan that will achieve a desired goal state in the environment.

\subsection{Concept Acquisition for ITL}
\label{sec:concept-space}
With an understanding of how indexical comprehension connects language with perceptions and actions and how tasks are executed, we can begin to define the concept learning problem. Our main question is this - where does the conceptual knowledge in semantic memory (in Figure \ref{fig:reperesntation-indexical}) come from? We study how this knowledge is acquired through interactions with an intelligent trainer who demonstrates relevant concepts by structuring the learner's environment. In Soar, episodic memory stores contextual experiences while the semantic memory stores general, context-independent facts. Our approach uses supervision from an intelligent trainer to group contextual experiences together. An analogical generalization process distills the common elements in grouped contextual experience. This process can be seen as mediating knowledge in Soar's episodic and semantic memories.

To develop our ideas further, we focus on learning three kinds of concepts. These concepts are crucial for ELP and ITL. \textbf{Visual} concepts correspond to perceptual attributes of objects and include colors and shapes. They provide meaning to nouns and adjectives in the linguistic input. \textbf{Spatial} concepts correspond to configuration of objects and provide grounding to prepositional phrases in the linguistic input. \textbf{Action} concepts correspond to temporal changes in object configurations and provide grounding to verb phrases.

\subsection{Curriculum of Guided Participation}
\label{sec:curriculum}
We introduce a novel interactive process for training \textsc{Aileen} to recognize and use novel concepts - \emph{guided participation}. Guided participation sequences and presents lessons - conjoint stimuli (world and language) - to \textsc{Aileen}. A lesson consists of a scenario setup in \textsc{Aileen}'s world and an interaction with \textsc{Aileen}. A scenario can be a static scene when training visual and spatial concepts or a sequence of scenes when training an action concept. An interaction has a linguistic component (\emph{content}) and a non-linguistic component (\emph{signal}). The \texttt{signal} component of instruction guides reasoning in \textsc{Aileen} and determines how it processes and responds to the content. Currently, \textsc{Aileen} can interpret and process the following types of signals:
\begin{enumerate}
     \item \texttt{inform}: \textsc{Aileen} performs active learning. It uses all its available knowledge to process the content through indexical comprehension (Section \ref{sec:icl}). If failures occur, \textsc{Aileen} creates a learning goal for itself. In this goal, it uses the current scenario to generate a concrete example of the concept described in the content. This example is sent to its concept memory. If no failure occurs, \textsc{Aileen} does not learn from the example. \textsc{Aileen} learning is deliberate; it evaluates the applicability of its current knowledge in processing the linguistic content. It learns only when the current knowledge isn't applicable, and consequently, \textsc{Aileen} accumulates the minimum number of examples necessary to correctly comprehend the content in its lessons.

    \item \texttt{verify}: \textsc{Aileen} analyzes the content through indexical comprehension and determines if the content refers to specific objects, spatial relationships, or actions in the accompanying scenario. If \textsc{Aileen} lacks knowledge to complete verification, \textsc{Aileen} indicates a failure to the instructor.

    \item \texttt{react}: This signal is defined only when the linguistic content contains a reference to an action. \textsc{Aileen} uses its knowledge to produce an action instantiation. Upon instantiation, \textsc{Aileen} determines a goal state in the environment and then plans, a sequence of actions to achieve the goal state. This sequence of actions is executed in the environment.
\end{enumerate}
Incorporating these variations in how \textsc{Aileen} responds to the linguistic content in a lesson enables flexible interactive learning. A trainer can evaluate the current state of knowledge in \textsc{Aileen} by assigning it verify and react lessons. While the verify lesson tests if \textsc{Aileen} can recognize a concept in the world, the \textsc{react} lesson tests if \textsc{Aileen} can use a known concept to guide its own behavior in the environment. Observations of failures helps the trainer in structuring inform lessons that guide \textsc{Aileen}'s learning. In an inform lesson, \textsc{Aileen} evaluates its own learning and only adds examples when necessary. Such learning strategy distributes the onus of learning between both participants. Lessons can be structured in a flexible, reactive way in real human-robot training scenarios.

\subsection{Desiderata for a Concept Memory}
\label{sec:desiderata}
We propose the following desiderata for a concept memory, which differs from previous approaches \citep{langley1987machine} due to our emphasis on embedding it within a larger task, in this case ELP and ITL:
\begin{enumerate}[label=D\arabic*, start=0]
    \item\label{d:0} Is (a) architecturally integrated and (b) uses relational representations.
    \item\label{d:1} Can represent and learn a diverse types of concepts. In particular, for \textsc{Aileen}, the concept memory must be able to learn visual concepts, spatial concepts, and action concepts.
    \item\label{d:2} Learn from exemplars acquired through experience in the environment. \textsc{Aileen} is taught through lessons that have two stimuli - a scenario and linguistic content that describes it.
    \item\label{d:3} Enable incremental accumulation of knowledge. Interactive learning is a distinctive learning approach in which behavior is intertwined with learning. It has been previously argued that interleaving behavior and learning splits the onus of learning between the instructor and the learner such that the instructor can observe the learner's behavior and provide more examples/instruction if necessary
    \item\label{d:4} Facilitate diverse reasoning over definitions of concepts.
    \begin{enumerate}
        \item Evaluate existence of a concept in the current environment, including its typicality. This enables recognizing a concept in the environment.
        \item Envision a concept by instantiating it in the current environment. This enables action in the environment.
        \item Evaluate the quality of concept definitions. This enables active learning - if the quality of a concept is poor, more examples can be added to improve it.
    \end{enumerate}
    \item\label{d:5} Learn from little supervision as realistically humans cannot provide a lot of examples.
\end{enumerate}

\section{Concept Memory}
\label{sec:concept-memory}
Concept learning in \textsc{Aileen} begins with a failure during Indexical comprehension in an inform lesson. Assume that \textsc{Aileen} does not know the meaning of \emph{red}, i.e, it does not know that \emph{red} implies the percept \texttt{CVRed} in the object description. When attempting to ground the phrase \emph{red cylinder} in our example, Indexical comprehension will fail when it tries to look-up the meaning of the word \emph{red} in its semantic memory. As in \textsc{Rosie}, a failure (or an impasse) in \textsc{Aileen} is an opportunity to learn. Learning occurs through interactions with a novel concept memory in addition to Soar's semantic memory. Similarly to Soar's dLTM, the concept memory is accessed by placing commands in a working memory buffer (a specific sub-graph). The concept memory interface has $4$ commands: \texttt{create}, \texttt{store}, \texttt{query}, and \texttt{project}. Of these, \texttt{store} and \texttt{query} are common with other Soar dLTMs. \texttt{create} and \texttt{project} are novel and explained in the following sections.

\textsc{Aileen}'s concept memory is built on two models of cognitive processes: SME \citep{forbus2017extending} and SAGE \citep{mclure2015extending} and can learn visual, spatial, and action concepts (desiderata \ref{d:0}). Below we describe how each function of concept memory is built with these models. The current implementation of the memory represents knowledge as predicate calculus statements or \emph{facts}, we have implemented methods that automatically converts Soar's object-oriented graph description to a list of facts when needed. Example translations from Soar's working memory graph to predicate calculus statements are shown in Table \ref{tab:rep-pred}. Visual and spatial learning requires generating facts from the current scene. Examples for action learning are provided through a demonstration which is automatically encoded in Soar's episodic memory. An episodic trace of facts is extracted from the episodic memory (shown in Table \ref{tab:rep-pred}). We will rely on examples in Table \ref{tab:rep-pred} for illustrating the operation of the concept memory in the remainder of this section. We have summarized various terms and parameters used in analogical processing in Table \ref{tab:params}.

\begin{table}[]
\caption{Predicate calculus representation for the world scene in Figure \ref{fig:architecture} corresponding to Soar's working memory graph in Figure \ref{fig:reperesntation-indexical}. \texttt{CVCyl} is short for the \texttt{CVCylinder} symbol and \texttt{H} for that \texttt{holdsIn} predicate that encodes which predicates hold in which episodic timepoint.}
\label{tab:rep-pred}
\small
\ttfamily
\begin{tabular}{@{}lllll@{}}
\toprule
\multicolumn{2}{l}{\textnormal{\textbf{Current world scene}}} & \multicolumn{3}{l}{\textnormal{\textbf{Episodic trace}}}                                                                            \\ \midrule
objects                 & relations     & T0                           & T1                                          & T2                     \\ \midrule
(isa o1 CVBlue)         & (e o1 o2)     & (H T0 (dc o1 o2)) & (H T1 (held O1))      &  (H T2 (w o1 o2)) \\
(isa o1 CVCone)         & (dc o1 o2)    & (H T0 (e o1 o2))   & ... & ...                    \\
(isa o2 CVRed)          & (w o2 o1)     & ...                          & ...                         & (final T2 T1)          \\
(isa o2 CVCyl)     & (dc o2 o1)    & (isa T0 start)               & (after T1 T0)               & (after T2 T1)       \\
\bottomrule
\end{tabular}
\end{table}

\subsection{Creation and Storage}
\begin{figure}[t]
    \centering
    \includegraphics[width=0.4\textwidth]{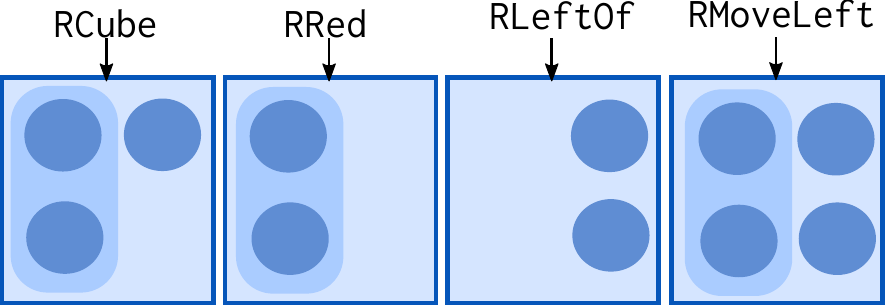}
    \qquad\begin{tabular}[b]{ll}
    \toprule
    Facts & P \\
    \midrule
    \texttt{(isa (GenEntFn 0 RRedMt) RRed)} & 1.0  \\
    \texttt{(isa (GenEntFn 0 RRedMt) CVRed)} & 1.0  \\
    \texttt{(isa (GenEntFn 0 RRedMt) CVCube)} & 0.5  \\
    \texttt{(isa (GenEntFn 0 RRedMt) CVCylinder)} & 0.5  \\
    \bottomrule
    \end{tabular}
    \caption{(left) SAGE maintains a generalization context for each concept. For each example (circle) of a concept, it is either added to a generalization (rounded rectangle) or maintained as an independent example for the concept. (right) Facts and their probabilities in generalization context for \texttt{RRed}}
    \label{fig:gcontext}
  \end{figure}

When \textsc{Aileen} identifies a new concept in linguistic content (word \emph{red}), it creates a new symbol \texttt{RRed}. This new symbol in incorporated in a map in Soar's semantic memory and is passed on to the concept memory for creation of a new concept via the \texttt{create} command. The concept memory creates a new reasoning symbol as well as a new generalization context (shown in Figure \ref{fig:gcontext}). A generalization context is an accumulation of concrete experiences with a concept. Each generalization context is a set of individual examples and generalizations.

\begin{table}[b]
    \small
    \centering
    \caption{Terms used in analogical processing, their definitions, and values in \textsc{Aileen}'s concept memory}
    \label{tab:params}
    \begin{tabular}{lp{10.8cm}l}
    \toprule
         \textbf{Term} & \multicolumn{2}{l}{\textbf{Definition}}\\
         Similarity & \multicolumn{2}{p{12cm}}{The score representing the quality of an analogical match, degree of overlap} \\
         Correspondence & A one-to-one alignment between the compared representations & \\
         Candidate Inference & Inferences resulting from the correspondences of the analogy& \\
         \midrule
         \textbf{Threshold} & \textbf{Definition}  & \textbf{Value}\\
         Assimilation & Score required to include a new example into a generalization instead of storing it as an example & 0.01 \\
         Probability & Only facts exceeding this value are considered part of the concept. & 0.6  \\
         Match & Score required to consider that an inference is applicable in a given scene & 0.75 \\
         \bottomrule
    \end{tabular}
\end{table}

After creating a new concept, Soar stores an example in the concept memory. The command \texttt{\{store: [(isa o2 CVRed) (isa o2 CVCylinder) (isa o2 RRed)], concept: RRed\}} stores that the object \texttt{o2} in the world is an example of the concept \texttt{RRed}. This example A is stored in the \texttt{RRed} generalization context as is - as a set of facts. Assume that at a later time, Soar sends another example B of \texttt{RRed} concept through the command \texttt{\{store: [(isa o3 CVRed) (isa o3 CVCube) (isa o3 RRed)], concept: RRed\}}. The concept memory adds the new example to the \texttt{RRed} generalization context by these two computational steps:
\begin{enumerate}
    \item SME performs an analogical match between the two examples. The result of analogical matching has two components: a correspondence set and a similarity score. A correspondence set contains alignment of each fact in one example with at most one fact from other. The similarity score indicates the degree of overlap between the two representations. In the two examples A and B, there are two corresponding facts: \texttt{(isa o2 CVRed)} aligns with \texttt{(isa o3 CVRed)} and \texttt{(isa o2 RRed)} aligns with \texttt{(isa o3 RRed)}. If the similarity score exceeds an \emph{assimilation threshold} (Table \ref{tab:params}), SAGE continues to the next step to create a generalization.
    \item SAGE assimilates the two examples A and B into a generalization (e.g. Figure \ref{fig:gcontext}, right). It :
    \begin{enumerate}
        \item Uses the correspondence to create abstract entities. In the two examples provided, \texttt{(isa o2 RRed)} aligns with \texttt{(isa o3 RRed)} and \texttt{(isa o2 CVRed)} with \texttt{(isa o3 CVRed)}. Therefore, identifiers \texttt{o2} and \texttt{o3} can be replaced with an abstract entity \texttt{(GenEntFn 0 RRedMt)}.
        \item Maintains a probability that a fact belongs in the generalization. Because \texttt{(isa (GenEntFN 0 RRedMT) RRed)} and \texttt{(isa (GenEntFn 0 RRedMT) CVRed)} are common in both examples, they are assigned a probability of $1$. Other facts are not in the correspondences and appear in $1$ of the $2$ examples in the generalization resulting in a probability of $0.5$. Each time a new example is added to this generalization, the probabilities will be updated the reflect the number of examples for which the facts were aligned with each other.
    \end{enumerate}
\end{enumerate}
Upon storage in a generalization context, a generalization becomes available for matching and possible assimilation with future examples enabling incremental (\ref{d:3}), example-driven (\ref{d:2}) learning.

\subsection{Query}
\begin{figure}
    \centering
    \includegraphics{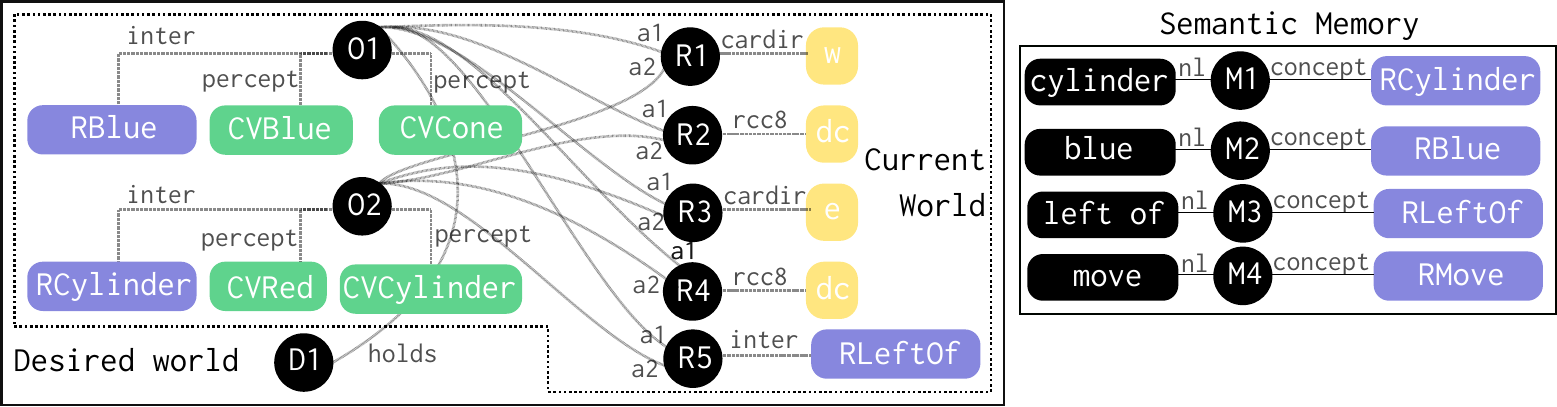}
    \caption{Working memory graph corresponding to scene in Figure \ref{fig:architecture} now enhanced with concept symbols (blue). Each concept symbol refers to a generalization context in the concept memory. The graph is enhanced based on inferences supported by analogical processing.}
    \label{fig:enhanced-wm-graph)}
\end{figure}

During indexical comprehension, \textsc{Aileen} evaluates if a known concept exists in the current world through the \texttt{query} command. Assume that in an example scene with two objects, indexical comprehension attempts to find the one that is referred to by \emph{red} through \texttt{\{query: \{scene: [(isa o4 CVRed) (isa o4 CVBox) (isa o5 CVGreen) (isa o2 CVCylinder)], pattern: (isa ?o RRed)\}\}}. In response to this command, the concept memory evaluates if it has enough evidence in the generalization context for \texttt{RRed} to infer \texttt{(isa o2 RRed)}. The concept memory performs this inference through the following computations.
\begin{enumerate}
    \item SME generates a set of candidate inferences. It matches the \texttt{scene} with the generalization in Figure \ref{fig:gcontext} (right). This match results in a correspondence between the facts \texttt{(isa o4 CVRed)} in \texttt{scene}) and \texttt{(isa (GenEntFn 0 RRedMt) CVRed)}, which aligns \texttt{o4} with \texttt{(GenEntFn 0 RRedMt)}. Other facts that have arguments that align, but are not in the correspondences, are added to the set of candidate inferences. In our example, a candidate inference would be \texttt{(isa o4 RRed)}.
    \item \textsc{AILEEN} filters the candidate inferences based on the pattern in the query command. It removes all inferences that do not fit the pattern. If the list has an element, further support is calculated.
    \item \textsc{AILEEN} evaluates the support for inference by comparing the similarity score of the match to the \emph{match threshold}. That is, the more facts in the generalization that participate in the analogical match then it is more likely that the inference is valid.
\end{enumerate}
Through queries to the concept memory and resultant analogical inferences, the working memory graph (of the world in Figure \ref{fig:enhanced-wm-graph)}) is enhanced. This enhanced working memory graph supports indexical comprehension in Section \ref{sec:icl}. Note that the internal concept symbols in blue (such as \texttt{RBlue}) are generalization contexts in the concept memory that accumulate examples from training. Consequently, the `meaning' of the world \emph{blue} will evolve as more examples are accumulated.

\subsection{Projection}
In ITL, simply recognizing that an action has been demonstrated is insufficient, the agent must also be able to perform the action if directed (desiderata \ref{d:4}). One of the advantages of analogical generalization is that the same mechanism is used for recognition and projection. Consider the example scene Figure \ref{fig:architecture} in which the trainer asks \textsc{Aileen} to \emph{move the blue cone to the right of the red cylinder} using the \texttt{react} signal. Assume that \textsc{Aileen} has previously seen some other examples of this action that are stored in concept memory as episodic traces (an example is shown in \ref{tab:rep-pred}).

During indexical comprehension, \textsc{Aileen} performs queries to identify the \emph{blue cone}, \texttt{O1}, and \emph{red cylinder}, \texttt{O2}. Similarly, it maps the verb and the related preposition to \texttt{RMove} and \texttt{RRightOf}. To act, \textsc{Aileen} uses its concept memory to project the action through the command \texttt{\{project: \{trace: [(H T0 (dc o1 o2)) (H T0 (e o1 o2)) (isa AileenStartTime T0) ...], concept: RMove\}}. In response, the concept memory performs the following computations:

\begin{enumerate}
    \item SME generates a set of candidate inferences. SME to matches the current scene expressed as a trace against the generalization context of the action \texttt{RMove}. SME generates all the candidate inferences that symbolically describe the next states of the action concept.
    \item \textsc{Aileen} filters the candidate inferences to determine which apply in the immediate next state (shown in Figure \ref{fig:project}). For example, the trace in the project command contains episode \texttt{T0} as the \texttt{AileenStartTime}. The filter computation will select facts that are expected to be held in \texttt{T1} and that the \texttt{(after T1 T0)} holds.
\end{enumerate}

\begin{figure}[h!]
    \centering
    \texttt{(H (:skolem (GenEntFn 0 0 rMoveMt)) (held O1)} \\
    \texttt{(after (:skolem (GenEntFn 0 0 rMoveMt)) T0)}
    \caption{Candidate inferences indicate that the next state of the move action is to hold object \texttt{O5}. Skolem terms are generated by SME to indicate that the candidate inference refers to an entity from the concept for which there is no correspondence in the scene. In this case, the skolem represents the next temporal state of the action as denoted by the \texttt{after} relation.}
    \label{fig:project}
  \end{figure}

This retrieval is accepted by \textsc{Aileen} to be next desired state it must try to achieve in the environment. As previously described, \textsc{Aileen} plans using its pre-encoded actions models and iterative deepening search. The search results in \texttt{pick-up(O1)}. After executing a \texttt{pick-up} action, \textsc{Aileen} invokes projection again to determine if \texttt{RMove} requires more steps. In this case, it does, and the candidate inferences specify that \texttt{O1} should be located to the \texttt{w} of \texttt{O2} and they should be topologically disjoint. Further, these candidate inferences indicate that this is the last step in the action, and therefore \textsc{Aileen} marks the action as completed after executing it.

\section{Evaluation}
\label{sec:evaluation}
As per desiderata \ref{d:1}, a concept memory must be able to learn diverse types of concepts. We demonstrate this capability by demonstrating learning of visual, spatial, and action concepts. The concepts are taught through lessons in a curriculum of guided participation (Section \ref{sec:curriculum}) demonstrating that concepts can be learned from grounded exemplars (\ref{d:2}). The experiments emulate aspects of ITL with humans where concepts are introduced incrementally during behavior (\ref{d:3}).

\begin{figure}[]
\center{\includegraphics[width=0.9\linewidth]{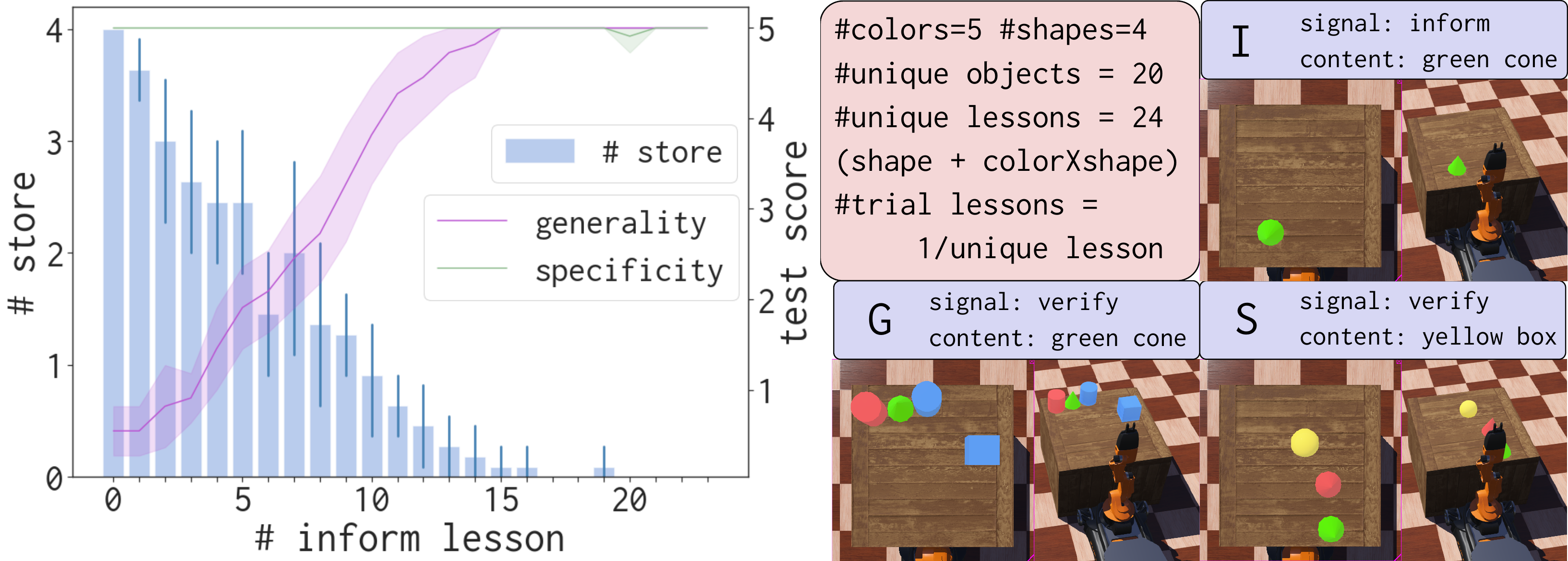}}
        \caption{\label{fig:object-res} (left) Learning curve for visual concepts averaged from $10$ trials. A trial includes lessons from $5$ colors and $4$ and shapes $= 20$ unique objects. Lessons include reference only to shape and color and shape. (right) Examples of an \emph{inform} lesson (I) and generality (G) and specificity (S) exam lessons. The blue bars show the average number of \texttt{create} or \texttt{store} commands executed in the concept memory. The pink link shows average score on the generality exam and the green bar at the top shows the average score on the specificity exam.}
\end{figure}

\begin{figure}[]
\center{\includegraphics[width=0.9\linewidth]{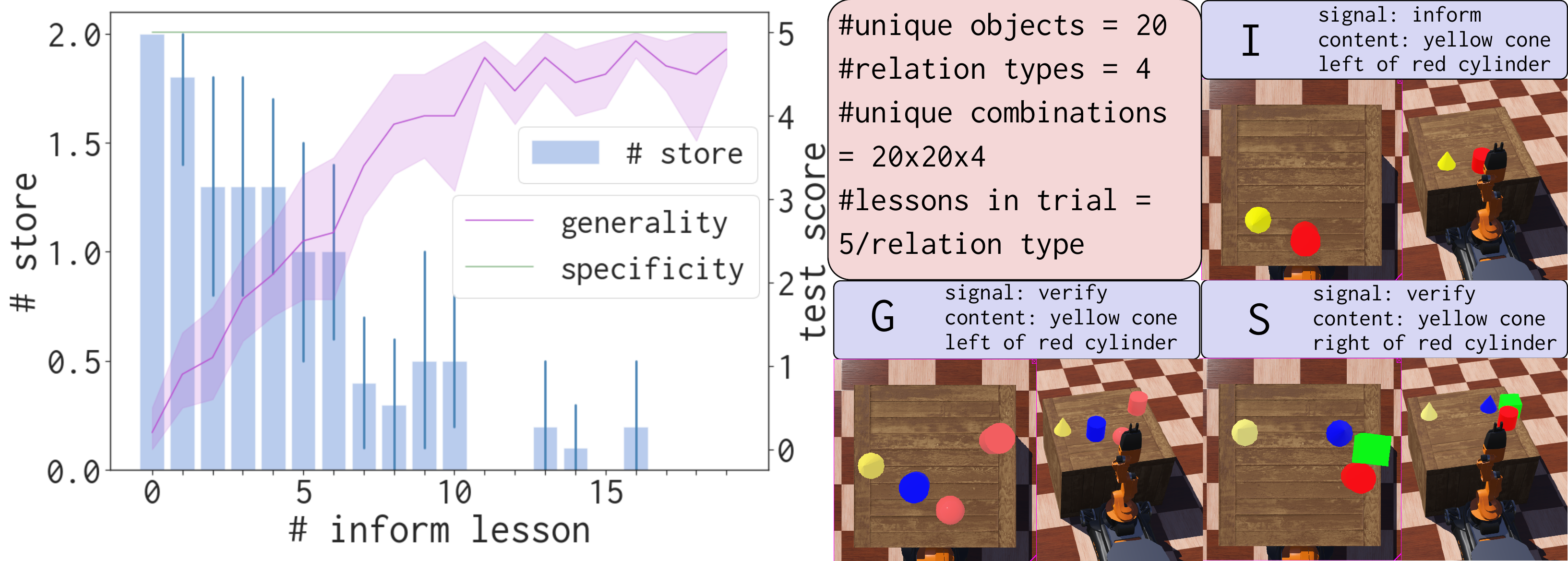}}
        \caption{\label{fig:relation-res} (left) Learning curve for spatial concepts averaged from $10$ trials. A trial includes lessons about $4$ types of binary relations defined over $20$ unique objects. (right) Examples of an \emph{inform} lesson (I) and generality (G) and specificity (S) exam lessons. The blue bars show the average number of \texttt{create} or \texttt{store} commands executed in the concept memory. The pink link shows average score on the generality exam and the green bar at the top shows the average score on the specificity exam.}
\end{figure}

\begin{figure}[]
\center{\includegraphics[width=\linewidth]{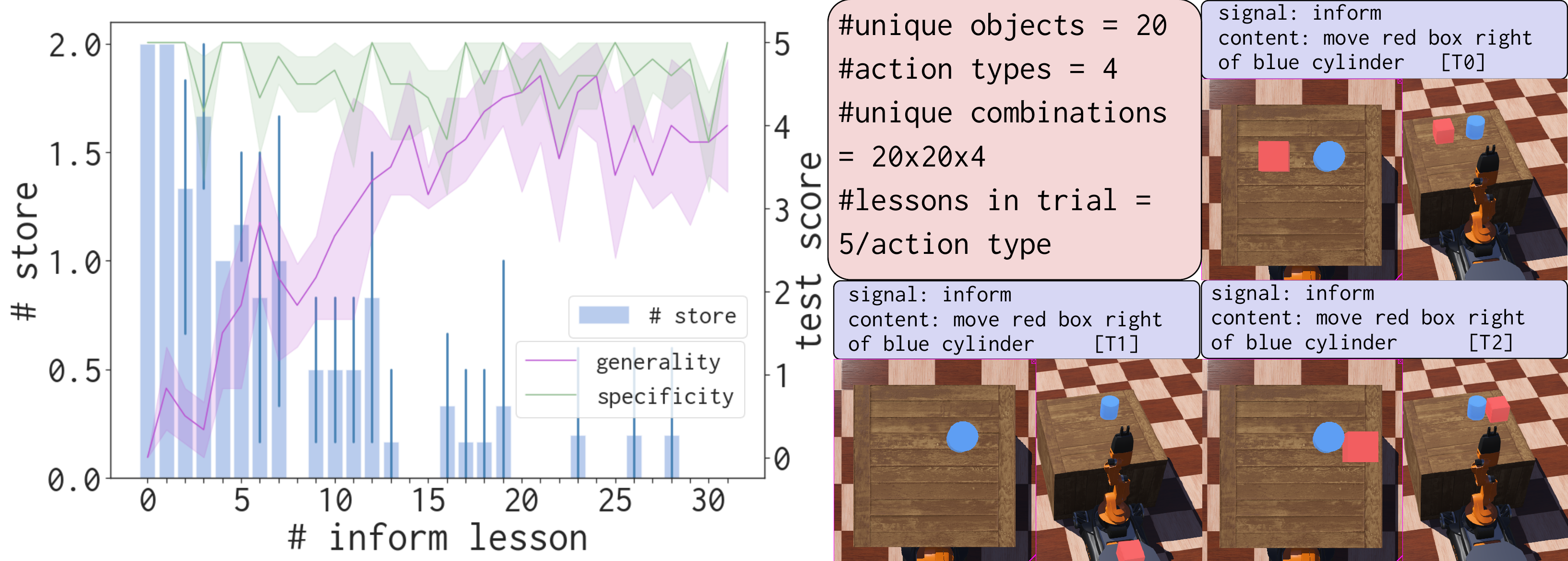}}
        \caption{\label{fig:action-res} (left) Learning curve for action concepts averaged from $5$ trials. A trial includes lessons about $1$ verb \emph{move} with $4$ different relations and two objects chosen from $20$ unique objects. The blue bars show the average number of \texttt{create} or \texttt{store} commands executed in the concept memory. The pink link shows average score on the generality exam and the green bar at the top shows the average score on the specificity exam. (right) A demonstration.}
\end{figure}

\paragraph{Method}
We performed separate learning experiments for visual, spatial, and action concepts. We leverage the lessons of guided participation in the design of our experimental trials. Each trial is a sequence of \emph{inform} lessons. In an \emph{inform} lesson, a concept is randomly selected from a pre-determined set and shown to \textsc{Aileen} accompanied with linguistic content describing the concept. The lesson is simplified, i.e, there are no distractor objects (examples are shown in Figures \ref{fig:object-res}, \ref{fig:relation-res}, \& \ref{fig:action-res}). The lesson is presented to \textsc{Aileen} and we record the number of store requests it makes to the concept memory. Recall that \textsc{Aileen} learns actively; i.e, it deliberately evaluates if it can understand the linguistic content with its current knowledge and stores examples only when necessary. The number of store requests made highlight the impact of such active learning.

Additionally, to measure generality and correctness, we test \textsc{Aileen} knowledge after every \emph{inform} lesson through two exams: generality and specificity (examples are shown in Figures \ref{fig:object-res}, \ref{fig:relation-res}, \& \ref{fig:action-res}). Both exams are made up of $5$ \emph{verify} lessons that are randomly selected at the beginning of the trial. As \textsc{Aileen} learns, the scores on these test demonstrate how well \textsc{Aileen} can apply what it has learned until now. In the generality lessons, \textsc{Aileen} is asked to verify if the concept in the linguistic input exists on the scene. If \textsc{Aileen} returns with a success status, it is given a score of $1$ and $0$ otherwise. In the specificity exam, \textsc{Aileen} is asked to verify the existence of a concept, however, the scenario does not contain the concept that is referred to in the linguistic content. If \textsc{Aileen} returns with a failed status, it is given a score of $1$ and $0$ otherwise. Both types of exam lessons have $0-3$ distractor objects introduced on the scene to evaluate if existence of noise impacts the application of conceptual knowledge.

\paragraph{Results}
Figure \ref{fig:object-res} illustrates visual concept learning. \textsc{Aileen} begins without any knowledge of any concept. As two concepts (\emph{green} and \emph{cone}) are introduced in the first lesson, it provides several store commands to its concept memory (shown in blue bars). The number of commands reduce as the training progresses. As is expected, the score on the generality exam is very low because \textsc{Aileen} doesn't know any concepts. However, this score grows very quickly with training eventually reaching perfect performance at lesson $15$. The score on the specificity exam starts at $5$, this is to be expected as well. This is because if a concept is unknown, \textsc{Aileen} cannot recognize it on the scene. However, as the trial progress we see that this score doesn't drop. This indicates that conceptual knowledge of one concept doesn't bleed into others. Note that the exams have distractor objects while learning occurred without any distractors - good scores on these exams demonstrate the strength of relational representations implemented in \textsc{Aileen}. Finally, \textsc{Aileen} learns from very few examples indicated that such learning systems can learn online with human trainers.

Figure \ref{fig:relation-res} illustrates spatial concept learning (commenced after all visual concepts are already known). Spatial relationships are defined between two objects each of which can be $1/20$ possible in the domain. Concrete examples include irrelevant information (e.g., \emph{left of}" does not depend on visual properties of the objects). Despite this large complex learning space, learning is quick and spatial concepts can be learned with few examples. These results demonstrate the strength of analogical generalization over relational representations. An interesting observation is that generality scores do not converge to $5$ as in visual concept learning. A further analysis revealed that in noisy scenes when the trainer places several distractors on the scene, sometimes the objects move because they are placed too close and the environment has physics built into it. The movement causes objects to move from the intended configuration leading to apparent error in \textsc{Aileen}'s performance. This is a problem with our experimental framework. The learning itself is robust as demonstrated by number of store commands in the trial which reduce to $0$ at the end.

Figure \ref{fig:action-res} illustrates action learning (commenced after all visual and spatial concepts have been learned). Actions are generated through the template \emph{move <object reference 1> <relation> <object reference 2>}. Similarly to spatial concepts, the learning space is very large and complex. When \textsc{Aileen} asks, it is provided a demonstration of action performance as shown in Figure \ref{fig:action-res} (\texttt{T0}, \texttt{T1}, \texttt{T2}). \textsc{Aileen} stores the demonstration trace in its episodic memory. For storing an example in the concept memory, information in Soar's episodic memory is translated into an episodic trace as shown Table \ref{tab:rep-pred}. Similarly to visual and spatial learning, inform lessons with simplified scene are used to teach a concept. Exams made up of positive and negative verify lessons are used to evaluate learning. As we see in Figure \ref{fig:action-res}, \textsc{Aileen} can quickly learn action concepts. Errors towards the later part of the experimental trial occur for the same reason we identified in spatial learning.

\paragraph{Task Demonstration} After visual, spatial, and action concepts were taught, we used a react lesson to see if \textsc{Aileen} could perform the actions such as \emph{move the red cylinder right of the blue box}. \textsc{Aileen} successfully use analogy-driven planning. After every action, it queried the concept memory to find the next desired state that it should try to achieve in the environment using the project command. After retrieving the next state, it applied the models of actions to determine which action will achieve the desired state. Upon finding an action, it applied it to the environment. After $2$ successive projections and search steps, if achieved the goal. This is extremely encouraging - \textsc{Aileen} could learn to perform an action by learning through demonstrations. Further, this evaluation establishes that the concept represented explored in this paper not only support interactive learning of concepts, they can be applied for recognition as well as action.

\section{Related Work}
\label{sec:related-work}
Diverse disciplines in AI have proposed approaches for concept learning from examples however, not all approaches can be integrated in a CMC architecture. We use the desiderata defined in Section \ref{sec:desiderata} to evaluate the utility of various concept learning approaches. The vast majority study the problem in isolation and consider only flat representations violating the desiderata \ref{d:0}. ML-based classification approaches are designed for limited types of concepts, violating desiderata \ref{d:1}, and require a large number of examples, violating desiderata D5, which are added in batch-mode, violating desiderata D3. More specifically, while EBL and
Inductive logic programming \citep{muggleton1994inductive} can learn from few datapoints, they require fully-specified domain theory violating desiderata \ref{d:2}. Bayesian concept learning \cite{tenenbaum1999bayesian} uses propositional representations, violating \ref{d:0}, and each demonstration has focused on a single type of concept, violating
\ref{d:1}.

There are a few cognitive systems' approaches to the concept learning problem that aim toward the desiderata that we delineated in Section \ref{sec:icl}. In the late $1980$s - early $1990$s, there was a concerted effort to align machine learning and cognitive science around concept formation \citep{fisher1987knowledge}. For example, \textsc{Labyrinth} \citep{thompson1991concept} creates clusters of examples, summary descriptions, and a hierarchical organization of concepts using a sequence of structure examples. \textsc{COBWeb3} \citep{fisher1987knowledge} incorporates numeric attributes and provides a probabilistic definition differences between concepts. Building off these ideas, \textsc{Trestle} \citep{maclellan2015trestle} learns concepts that include structural, relational, and numerical information. Our work can be seen as a significant step along this direction. First, the concept memory proposed here leverages the computational models of analogical processing that have been shown to emulate analogical reasoning in humans. Second, we place the concept learning problem within the larger problems of ELP and ITL in a cognitive architecture context. We demonstrate not only concept formation but also how learned concepts are applied for recognition, scene understanding, and action reasoning. By integrating with vision techniques, we demonstrate one way in which concept formation is tied to sensing.

Another thread of work in the cognitive system's community that we build upon is that of analogical learning and problem-solving. Early analogical problem-solving systems include Cascade \citep{vanlehn1991modeling}, Prodigy \citep{veloso1995integrating}, and Eureka \citep{jones2005constrained}. They typically used analogy in two ways: (1) as analogical search control knowledge where previous examples were used to guide the selection of which problem-solving operator to apply at any time, and (2) for the application of example-specific operators in new situations. \textsc{Aileen} differs in two important ways: (1) it relaxes the need for explicit goals further in its use of projection to specify the next subgoal of an action, and (2) it uses analogical generalization on top of analogical learning to remove extraneous knowledge from the concept.


\section{Discussion, Conclusions, and Future Work}
\label{sec:conclusions}
In this paper, we explored the design and evaluation of a novel concept memory for Soar (and other CMC cognitive architectures). The computations in the memory use models of analogical processing - SAGE and SME. This memory can be used to acquire new situated, concepts in interactive settings. The concepts learned are not only useful in ELP and recognition but also in task execution. While the results presented here are encouraging, the work described in this paper is only a small first step towards an architectural concept memory. We have only explored a functional integration of analogical processing in Soar. The memory has not be integrated into the architecture but is a separate module that Soar interacts with. There are significant differences between representations that Soar employs and those in the memory. For an efficient integration and a reactive performance that Soar has historically committed to, several engineering enhancements have to be made.

There are several avenues for extending this work. We are looking at three broad classes of research: disjunctive concepts, composable concepts, and expanded mixed-initiative learning. Disjunctive concepts arise from homographs (e.g., \emph{bow} in musical instrument versus \emph{bow} the part of a ship) as well as when the spatial calculi does not align with the concept or the functional aspects of the objects must be taken into account (e.g., a cup is \emph{under} a teapot when it is under the spigot, while a saucer is \emph{under} a cup when it is directly underneath). One of the promises of relational declarative representations of the form learned here is that they are composable. This isn't fully exploited for learning actions with spatial relations in them. Our approach ends up with different concepts for \texttt{move-left} and \texttt{move-above}. A better solution would be to have these in the same generalization such that \textsc{Aileen} would be able to respond to the command to \emph{move cube below cylinder} assuming it been taught a \emph{move} action previously along with the concepts for \emph{below}, \emph{cube}, and \emph{cylinder}. Another avenue is contextual application of concepts. For example, \emph{bigger box} requires comparison between existing objects. Finally a cognitive system should learn not only from a structured curriculum designed by an instructor but also in a semi-supervised fashion while performing tasks. In our context this means adding additional examples to concepts when they were used as part of a successful execution. This also means, when there are false positives that lead to incorrect execution, revising the learned concepts based on this knowledge. One approach from analogical generalization focuses on exploiting these near-misses with SAGE \citep{mclure2015extending}.

Inducing general conceptual knowledge from observations is a crucial capability of generally intelligent agents. The capability supports a variety of intelligent behavior such as operation in partially observable scenarios (where conceptual knowledge elaborates what is not seen), in language understanding (including ELP), in commonsense reasoning, as well in task execution. Analogical processing enables robust incremental induction from few examples and has been demonstrated as a key cognitive capability in humans. This paper explores how analogical processing can be integrated into the Soar cognitive architecture which is capable of flexible and contextual decision making and has been widely used to design complex intelligent agents. This paper paves way for an exciting exploration of new kinds of intelligent behavior enabled by analogical processing.

\section{Acknowledgements}
The authors thank Ken Forbus and Irina Rabkina for their support working with SME and SAGE. The authors appreciate Preeti Ramaraj and Charlie Ortiz for feedback and copyediting and John Laird for his suggestions on situating the claims and contributions. The anonymous reviewers for ACS 2020 provided very useful feedback on improving readability and highlighting connections to previous work. The work presented in this paper was supported in part by the DARPA GAILA program under award number HR00111990056 and an AFOSR grant on Levels of Learning, a sub-contract from University of Michigan (PI: John Laird, FA9550-18-1-0180).

The views, opinions and/or findings expressed are those of the authors' and should not be interpreted as representing the official views or policies of the Department of Defense, AFOSR, or the U.S. Government.

\setlength{\bibsep}{0pt}
{\parindent -10pt\leftskip 10pt\noindent
\bibliographystyle{cogsysapa}
\bibliography{references}
}

\end{document}